\documentclass[aps,prl,twocolumn,superscriptaddress]{revtex4-1}
\usepackage{amsfonts}
\usepackage{graphicx}
\usepackage{epstopdf}
\usepackage{epsfig}
\usepackage{amsmath}
\usepackage[normalem]{ulem}
\usepackage{color}
\usepackage{makecell}
\usepackage{array}
\usepackage{booktabs}
\usepackage{mathtools}
\usepackage{bm}
\usepackage{color}
\usepackage{array}
\usepackage{booktabs}
\usepackage{tabularx}

\usepackage{xr}

\definecolor{blue}{rgb}{0,0,1}

\newcolumntype{Y}{>{\centering\arraybackslash}X}

\newcommand{\Rmnum}[1]{\expandafter\@slowromancap\romannumeral #1@}
\makeatother

\begin{document}

\title{Machine learning prediction of critical transition and system collapse}

\date{\today}

\author{Ling-Wei Kong}
\affiliation{School of Electrical, Computer and Energy Engineering, Arizona State University, Tempe, AZ 85287, USA}

\author{Hua-Wei Fan}
\affiliation{School of Physics and Information Technology, Shaanxi Normal University, Xi'an 710062, China}

\author{Celso Grebogi}
\affiliation{Institute for Complex Systems and Mathematical Biology, School of Natural and Computing Sciences, King's College, University of Aberdeen, UK}

\author{Ying-Cheng Lai} \email{Ying-Cheng.Lai@asu.edu}
\affiliation{School of Electrical, Computer and Energy Engineering, Arizona State University, Tempe, AZ 85287, USA}
\affiliation{Department of Physics, Arizona State University,
Tempe, Arizona 85287, USA}

\begin{abstract}

	To predict a critical transition due to parameter drift without relying on model is an outstanding problem in nonlinear dynamics and applied fields. A closely related problem is to predict whether the system is already in or if the system will be in a transient state preceding its collapse. We develop a model free, machine learning based solution to both problems by exploiting reservoir computing to incorporate a parameter input channel. We demonstrate that, when the machine is trained in the normal functioning regime with a chaotic attractor (i.e., before the critical transition), the transition point can be predicted accurately. Remarkably, for a parameter drift through the critical point, the machine with the input parameter channel is able to predict not only that the system will be in a transient state, but also the average transient time before the final collapse\footnote{This paper was first submitted to {\it Physical Review Letters} on May 28, 2020.}.  

\end{abstract}

\maketitle

In nonlinear and complex dynamical systems, a catastrophic collapse is often
preceded by transient chaos. For example, in electrical power systems, 
voltage collapse can occur after the system enters into the state of transient 
chaos~\cite{DL:1999}. In ecology, slow parameter drift caused by environmental 
deterioration can induce a transition into transient chaos, after which species
extinction follows~\cite{McY:1994,HACFGLMPSZ:2018}. A common route to transient
chaos is a global bifurcation termed crisis~\cite{GOY:1983}, at which a chaotic 
attractor collides with its own basin boundary, is destroyed and becomes a 
chaotic transient. In the real world, the accurate system equations are often 
unknown but only measured time series are available. Model-free and 
data driven prediction of the critical transition and system collapse in 
advance of their occurrences, while the system is currently operating in a 
normal regime with a chaotic attractor, has been an outstanding problem. If 
the underlying equations of the system have a simple mathematical structure, 
e.g., with the velocity field consisting of power series or Fourier series 
terms only, then sparse optimization methods such as compressive sensing can 
be exploited to identify the system equations~\cite{WYLKG:2011,WLG:2016} and 
consequently to predict transitions. Here, we assume that our system does not
meet this condition and set out to develop a machine learning framework to 
predict the critical transition.    

A closely related problem is to determine if the system is already in a 
transient state - the question of ``how do you know you are in a transient?''. 
In nonlinear dynamics, this is one of the most difficult questions because 
the underlying system can be in a long transient in which all measurable 
physical quantities exhibit essentially the same behaviors as if the system 
were still in a sustained state with a chaotic attractor. Applying the 
traditional method of delay coordinate embedding~\cite{Takens:1980} to such 
a case would yield estimates of dynamical invariants such as the Lyapunov 
exponents and fractal dimensions, but it would give no indication that the 
system is already in a transient and so an eventual collapse is inevitable. 
Developing a model-free, purely data-driven predictive paradigm to address 
this problem is of considerable value to solving some of the most pressing 
problems in the modern time. Due to global warming and climate change, some 
natural systems may have already been in a transient state awaiting a 
catastrophic collapse to occur. A reliable determination at the present that 
the system has already passed the critical transition or a ``tipping'' point 
to a transient state would send a clear message to policy makers and the 
general public that actions must be taken immediately to avoid the otherwise 
inevitable catastrophic collapse.

\begin{figure}[ht!]
\centering
\includegraphics[width=\linewidth]{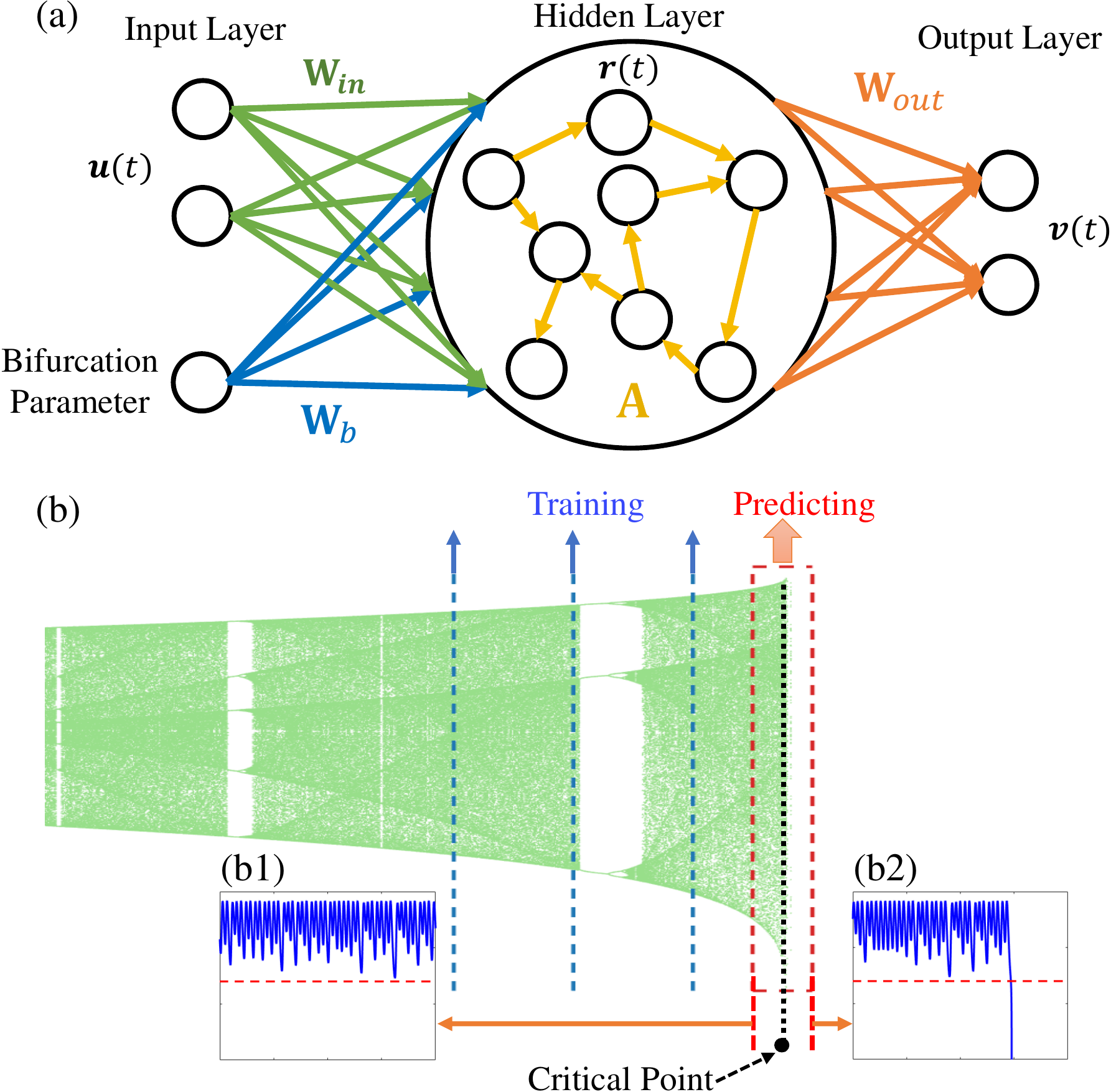}
\caption{
Modified reservoir computing scheme and illustration of the idea to predict 
critical transition and collapse. (a) Structure of the modified reservoir 
computing machine with an input parameter channel. The system consists of 
three layers: the input layer, the hidden layer and the output layer. Vectors 
$\mathbf{u}(t)$, $\mathbf{r}(t)$, and $\mathbf{v}(t)$ denote the set of input 
signals (measured time series from the dynamical variables of the target 
system), the dynamical state of the reservoir network (hidden layer), and the 
set of output signals representing the machine's prediction. The bifurcation 
parameter is designated as an additional input.
(b) The reservoir computing machine is trained with time series taken from a 
few parameter values in the green regime in which the system operates normally 
- indicated by the three vertical blue dashed lines. After training, prediction
can be made either for parameter values before the critical point (still in 
the green regime) or after, where in the former, the machine will predict 
that the system is safe (b1) but, importantly, for the latter, the machine 
will predict that the system will eventually collapse after transient chaos 
(b2).}
\label{fig:schematic}
\end{figure}

In this Letter, we develop a machine learning framework to predict critical
transition and transient chaos in nonlinear dynamical systems. We exploit 
reservoir computing, a class of recurrent neural networks~\cite{Jaeger:2001,
MNM:2002,JH:2004,MJ:2013}, a research area that has gained considerable 
momentum as a powerful paradigm for model-free prediction of nonlinear and 
chaotic dynamical systems~\cite{HSRFG:2015,LBMUCJ:2017,PLHGO:2017,LPHGBO:2017,
DBN:book,LHO:2018,PWFCHGO:2018,PHGLO:2018,Carroll:2018,NS:2018,ZP:2018,
WYGZS:2019,GPG:2019,JL:2019,VPHSGOK:2019,FJZWL:2020}. Our articulated 
reservoir computing structure differs from the conventional one in that we 
designate an additional input channel for the bifurcation parameter, as shown
in Fig.~\ref{fig:schematic}(a). The basic idea of our framework is explained
in Fig.~\ref{fig:schematic}(b), a schematic bifurcation 
diagram of a typical nonlinear system. In the green region, there is a chaotic 
attractor together with periodic windows, in which the system functioning is 
normal. A catastrophic bifurcation occurs at the critical parameter value 
defining the end of the green region, where the chaotic attractor is destroyed 
through a crisis transition. For a parameter value slightly beyond the critical
point, there is transient chaos leading to collapse. Suppose that, currently, 
the system operates in the normal regime. Given a certain amount of parameter 
drift, the two goals are: (i) to predict the transition point so as to 
determine whether the system will be in a transient chaotic regime heading to 
collapse and (ii) if yes, on average how long the system could survive, i.e., 
to predict the average lifetime of the chaotic transient. Because time series 
data from multiple parameter values are needed, it is necessary to specify the 
parameter value at which the data are taken - a task that can be accomplished 
by inputting the parameter value to all nodes in the underlying neural network. 
We demonstrate that our proposed machine learning framework can 
accomplish the two goals with three examples: an electrical power system 
susceptible to voltage collapse through transient chaos~\cite{DL:1999}, a 
three-species food chain model~\cite{McY:1994} in ecology in which a 
catastrophic transition leads to transient chaos and then species extinction, 
and the Kuramoto-Sivashinsky system~\cite{Kuramoto:1978,Sivashinsky:1980} in 
the regime of transient spatiotemporal chaos~\cite{HNZ:1986}. We show that, 
training a reservoir network of reasonable size, e.g., 1000 nodes, with time 
series data taken from three parameter values in the normal chaotic regime,
the machine is able to predict not only the collapse point but also transient 
chaos for parameter values beyond the critical point. A remarkable feature is 
that, after the critical transition, the probability distribution of the 
transient lifetime of the machine generated trajectories from an ensemble of 
initial conditions agrees with the true distribution, indicating that, for a 
given parameter drift into the transient chaos regime, the machine is able to 
predict, on average, how long the system can ``survive'' before collapse. 

For simplicity, we consider a catastrophic transition caused by variations 
in a single parameter. (Situations of multiple parameters can be treated 
with multiple parameter channels.) 
As shown in Fig.~\ref{fig:schematic}(a), the input layer maps the 
low-dimensional time series data $\mathbf{u}(t)$ into the high-dimensional 
hidden state $\mathbf{r}(t)$ through a matrix $\mathcal{W}_{in}$, and the 
output layer maps $\mathbf{r}(t)$ back into low-dimensional time series 
$\mathbf{v}(t)$ through another matrix $\mathcal{W}_{out}$. {\em The input 
parameter channel is connected to every node of the reservoir network} via 
the matrix $\mathcal{W}_{b}$. The reservoir network adjacency matrix 
$\mathcal{A}$ transfers information from the hidden states $\mathbf{r}(t)$ 
at $t$ to the next time step $\mathbf{r}(t+\Delta t)$. The dynamical 
evolution of the reservoir computing machine is described by
$\mathbf{r}(t+\Delta t) = (1-\alpha)\mathbf{r}(t) + \alpha\tanh{[\mathcal{A}\cdot\mathbf{r}(t)+\mathcal{W}_{in}\cdot\mathbf{u}(t)+k_b\mathcal{W}_b(\mathbf{b}+b_0)]}$ and $\mathbf{v}(t) = \mathcal{W}_{out}\cdot\mathbf{r}(t)$,
where $\alpha$ is the leakage parameter, $k_b$ and $b_0$ are two
hyperparameters determining the input of the bifurcation parameter into the
reservoir network. Training is administered to adjust the output matrix
$\mathcal{W}_{out}$ to minimize the difference between $\mathbf{v}(t)$ and
$\mathbf{u}(t+\Delta t)$, so that the reservoir can predict the evolution of
the target system into the future with input of the dynamical variables from
the past. The matrices $\mathcal{W}_{in}$, $\mathcal{W}_{b}$ and $\mathcal{A}$
are chosen {\em a priori} and fixed during the training and prediction phases
for efficiency.
(A more detailed description of the training and predicting processes
is provided in Sec.~I of Supplementary Information (SI)~\cite{SI}.)
We train the reservoir machine using time series from a few distinct
parameter values - all in the normal or safe regime where the system
still possesses a chaotic attractor, as shown in Fig.~\ref{fig:schematic}(b).
Because of the additional parameter input channel, the machine trained
with data at different parameter values will gain the ability to ``sense'' the
variation in the parameter and the associated changes in the time series
data from different parameter values. Such a well trained machine is a
high-dimensional representation of the original dynamical system. 
Finally, to predict with what parameter drift the system will exhibit transient
chaos and then collapse, we simply input the parameter value of interest into
the parameter input channel. Then, at each time step, we collect the one-step
prediction from the output layer $\mathbf{v}(t)$ and feed it back to the 
input layer $\mathbf{u}(t+\Delta t) = \mathbf{v}(t)$. Now the reservoir
machine in the predicting phase is a closed-loop dynamical system with one
constant external drive - the bifurcation parameter of interest. The 
machine in the predicting phase will be able to predict the system collapse
preceded by transient chaos for the input value of bifurcation parameter.

To achieve better performance, for each target system, we choose the seven key 
hyperparameters of the reservoir computing machine based on Bayesian 
optimization~\cite{GPG:2019}, which are the average degree and the spectral 
radius of the reservoir network, the scales of the input matrix 
$\mathcal{W}_{in}$, parameters $k_b$, $b_0$, and $\alpha$, as well as the 
regularization parameter used during the training of the output matrix 
$\mathcal{W}_{out}$.
(A more detailed description of these hyperparameters and their optimization 
can be found in Secs.~I and V of SI~\cite{SI}.) 

\begin{figure}[ht!]
\centering
\includegraphics[width=0.9\linewidth]{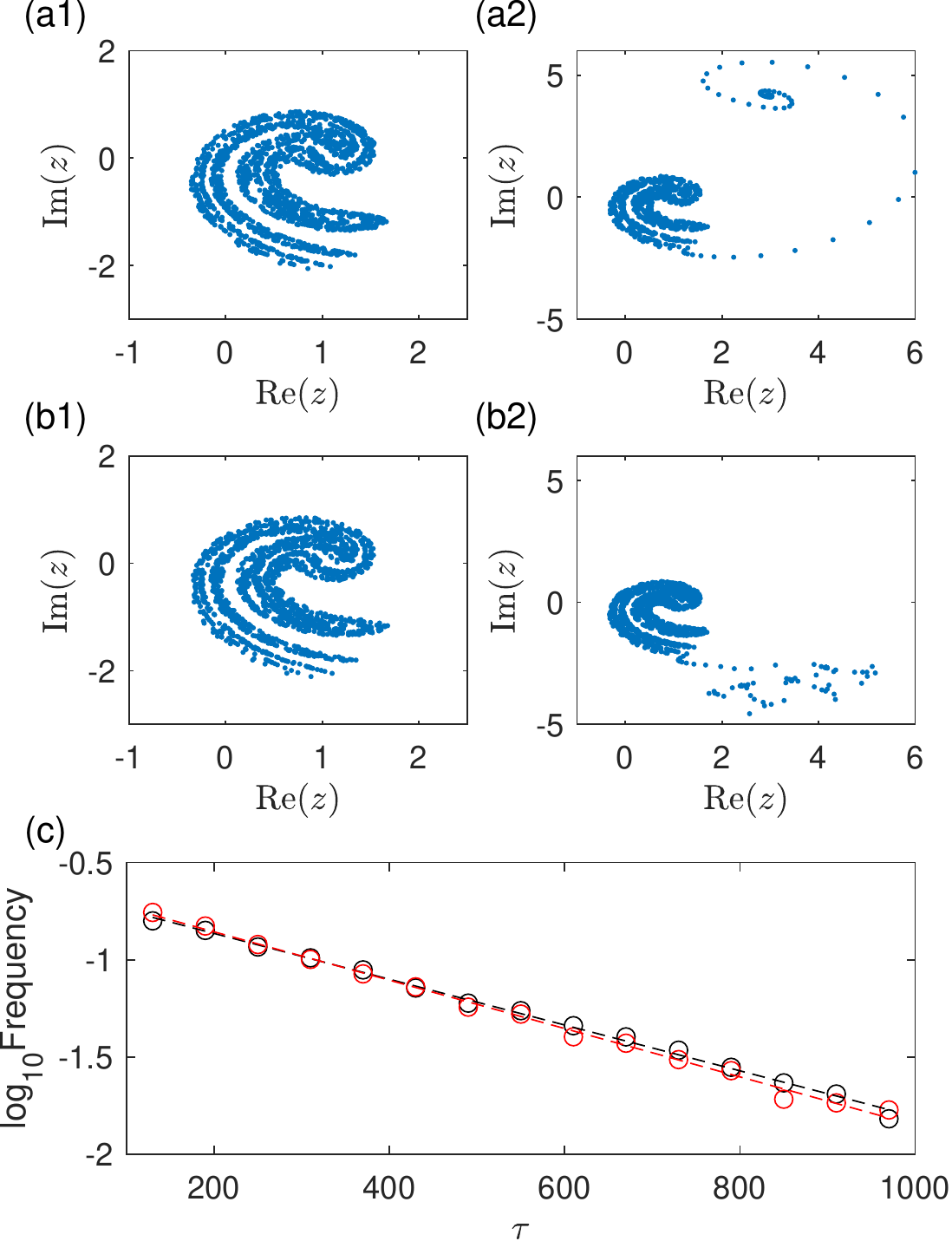}
\caption{ {Predicting transient chaos and collapse of dynamics
in a nonlinear optical cavity as described by the Ikeda map}.
(a1,a2) Typical behaviors of the system in the sustained and transient chaos
regimes, respectively, for $\mu=0.99<\mu_c$ and $\mu=1.01>\mu_c$. 
(b1,b2) Predicted behaviors by the reservoir machine for the same values of 
$\mu$ as in (a1,a2), respectively. The machine predicts correctly the 
system collapse for $\mu > \mu_c$. (c) Actual (black) and predicted (red) 
exponential transient lifetime distributions for $\mu=\mu_c+0.02$.}
\label{fig:Ikeda}
\end{figure}

To demonstrate that our machine learning approach represents a true advance
in nonlinear systems prediction, we present two examples for which the sparse 
optimization methods fail. In particular, the basic requirement of any sparse 
optimization technique for finding the system equations is {\em sparsity}: 
when the system equations are expanded into a power series or a Fourier series,
it must be that only a few terms are present so that the coefficient vectors 
to be determined from data are sparse~\cite{WYLKG:2011,WLG:2016}. 

The first example is the Ikeda map~\cite{Ikeda:1979,IDA:1980,HJM:1985} that 
describes the dynamics of a laser pulse propagating in a nonlinear cavity: 
$z_{n+1}= \mu+\gamma z_n \exp{ \left[ i\kappa-i\eta/(1+|z_n|^2) \right] }$,
where $z$ is a complex dynamical variable and the bifurcation parameter $\mu$ 
is the dimensionless laser input amplitude (more details in Sec.~IIA in 
SI~\cite{SI}). When the map functions are expanded into a power series, all 
combinations of the powers of the dynamical variables are necessary. The 
coefficient vectors for the dynamical variables are absolutely dense, 
representing an extreme type of violation of the sparsity condition. The 
system exhibits a boundary crisis~\cite{ISD:1998} at $\mu_c=1.0027$ and the 
dynamical behaviors for $\mu<\mu_c$ and $\mu>\mu_c$ are shown in
Figs.~\ref{fig:Ikeda}(a1) and ~\ref{fig:Ikeda}((a2), respectively. There is a 
chaotic attractor for $\mu<\mu_c$, and transient chaos leading to an escape 
of the system out of the previous operation region for $\mu > \mu_c$. We train 
the reservoir machine at $\mu=0.91,0.94,0.97$ - all in the chaotic attractor 
regime. For each $\mu$ value, training is done such that the reservoir machine 
is able to predict the exact state evolution of the original system for 
several Lyapunov time. More importantly, we ensure that the predicted 
trajectories land on the chaotic attractor for an arbitrarily long stretch 
of time. After training, we apply some parameter change $\Delta \mu$ and test, 
for each resulting parameter value, whether the predicted system state is a 
chaotic attractor or a chaotic transient. An exemplary pair of the predicted 
state for $\mu<\mu_c$ and $\mu>\mu_c$ are shown in Figs.~\ref{fig:Ikeda}(b1) 
and \ref{fig:Ikeda}(b2), respectively. It is remarkable that the reservoir 
machine is able to predict the collapse after a chaotic transient, as shown in 
Fig.~\ref{fig:Ikeda}(b2). Examining the prediction results for a set of 
systematically varied $\Delta \mu$ values enables determination of the 
critical bifurcation point, denoted as $\mu^*_{c}$. Averaging over 
1,000 independent random realizations of the reservoir configurations, we
obtain $\mu_c^*=1.00\pm0.01$, which agrees well with the actual value 
$\mu_c$. Our machine learning framework can also predict a basic
statistical characteristic of transient chaos: the lifetime distribution. To 
demonstrate this, we set the control parameter of the reservoir to be 
$\mu=\mu_c^*+0.02$ so that the system is in the transient chaos regime and 
the distribution of the transient lifetime is exponential. The reservoir 
system predicts correctly the exponential distribution, as shown in 
Fig.~\ref{fig:Ikeda}(c), where 50 stochastic realizations of the reservoir 
system and 400 random initial conditions for each machine realization are 
used. It can be seen that predicted distribution agrees well with the true
one, demonstrating the power of our reservoir computing scheme for predicting 
transient chaos and system escape (collapse).

\begin{figure}[ht!]
\centering
\includegraphics[width=0.9\linewidth]{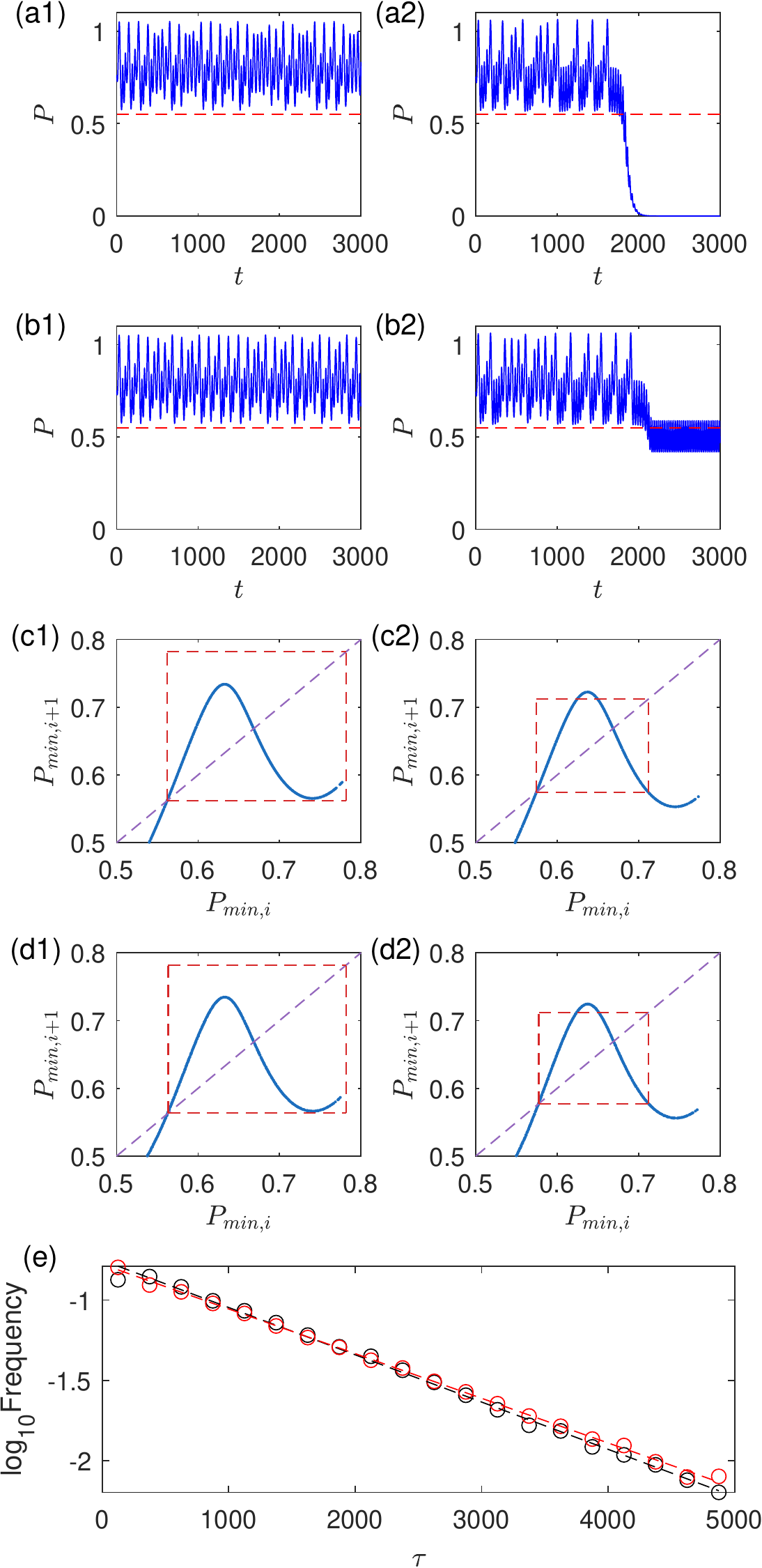}
\caption{ Predicting transient chaos and collapse of a food chain system.
(a1,a2) Typical time series of the predator density $P$ in the normal and 
transient regimes, respectively, for $K=0.997<K_c$ and $K=1.01>K_c$. 
(b1,b2) Predicted time series of $P$ for the same values of $K$ as in (a1,a2), 
respectively. The machine predicts correctly the sudden collapse for $K > K_c$.
(c1,c2) Return maps constructed from the local minima of $P(t)$ from the true 
time series, where the red dashed squares define an interval in which an 
invariant set exists: a chaotic attractor at $K = 0.997 < K_c$ (c1) or a
nonattracting chaotic set at $K = 1.01 >K_c$
due to the escaping region about the critical point leading to transient 
chaos (c2). (d1,d2) Predicted return maps for the same values of $K$ as in 
(c1,c2), respectively. (e) Actual (black) and predicted (red) transient 
lifetime distributions at $K=K_c+2\times10^{-4}$.}
\label{fig:FC}
\end{figure}

The second example is three-species food chain model~\cite{MY:1994} where, 
as the environmental capacity $K$ of the resource species is varied, a
catastrophic bifurcation and subsequent transient chaos leading to sudden
species extinction occurs. The system equations, when being expanded into 
a power series, have an infinite number of nontrivial terms, violating the
sparsity condition and rendering inapplicable any existing sparse optimization 
technique for data based discovery of the system equations  (detailed in 
Sec.~IIB in SI~\cite{SI}).
Figures~\ref{fig:FC}(a1) and (a2) show typical behaviors of the predator
density $P$ for $K<K_c$ and $K>K_c$ ($K_c = 0.99976$), where there is sustained
chaos in the former and transient chaos leading to species extinction in the
latter. We train the reservoir machine at $K=0.97$, $0.98$, and $0.99$ - all
in the sustained chaos regime. As shown in Figs.~\ref{fig:FC}(b1) and (b2),
the reservoir predicts correctly that, as the value of $K$ is increased, a
catastrophic crisis will occur. Averaging over an ensemble of 500 different 
reservoir realizations, we obtain the predicted critical transition point as 
$K^*_c=0.9997 \pm 4\times10^{-4}$, which agrees well with the actual
value $K_c=0.99976$.
Figure~\ref{fig:FC}(e) shows the predicted distribution of the transient 
lifetime obtained from 100 realizations of the reservoir machines, each with 
400 random initial conditions. The predicted average transient lifetime is 
about $1.35\times 10^3$, which agrees well with the true value 
($1.33\times 10^3$). All the transient lifetime are measured at 
values of $K$ that are $2\times10^{-4}$ beyond the systems' critical
points.

If predicting system collapse is viewed as a binary classification problem
(i.e., with or without collapse), it would be useful to train the neural
machine with data from both below and above the critical point. A difficulty
is that, beyond the critical point, the system will collapse after a transient
chaotic phase of random duration. Practically, it is infeasible to obtain
sufficient training data from the system in the post-critical regime. 
Moreover, our machine learning method can be used to assess the likelihood
of the occurrence of a crisis in the near future where no post-critical data 
are available.


Our results suggest that the reservoir machine trained with data from a
few distinct values of the bifurcation parameter represents a ``regression''
between the dynamical behavior of the target system and the bifurcation
parameter. The machine is able to make
statistically accurate predictions outside the training region. Since
training is done on as few as three different values of the bifurcation
parameter, the prediction of, e.g., the critical transition point, from
any individual reservoir realization will involve large errors. However,
the collective prediction from an ensemble of statistically independent
reservoir machines can be quite accurate. (A further analysis of the
reservoir performance is provided in Secs.~VI-VII in SI~\cite{SI}.)
In general, the prediction error would increase if the training parameter
values are further from the critical transition point, suggesting that
reservoir machines represent a low order approximation of the real dynamical
systems about the training points of the bifurcation parameter. This is a
natural and inevitable trade-off. 


In summary, we have articulated a parameter-aware scheme of reservoir 
computing to predict collapse as a result of parameter drift driving the 
system into transient chaos by designating an additional input channel to 
accommodate the bifurcation parameter, which is equivalent to introducing 
adjustable biases between the input and the hidden layers. With parameter
dependent training, all in the regime of sustained chaos, 
the reservoir machine acquires the ability to capture the variations 
in the ``climate'' of the target system, thereby gaining the power
to predict the system state for different parameter values. When a parameter
drift pushes the system through a critical point into a regime of transient 
chaos where collapse is inevitable, our design of machine learning is capable 
of accurate prediction of the critical value of the parameter, and of 
the statistical characteristics of transient chaos and the eventual collapse. 
These features are demonstrated using an electrical power system and a 
food-chain model in ecology. (An example of predicting transient 
spatiotemporal chaos in the Kuramoto-Sivashinsky system is given
in Sec.~III of SI~\cite{SI}.)

This work was supported by ONR under Grant No.~N00014-16-1-2828.


%
\end{document}